\RequirePackage{fix-cm}
\documentclass[smallextended]{for_arxiv}    

\smartqed  
\usepackage{graphicx}
\usepackage{amsmath}
\usepackage{amssymb}
\usepackage{bbm}
\usepackage{rotating}
\usepackage{color}
\usepackage{math}
\usepackage{booktabs}
\usepackage{url}
\usepackage{latexsym}
\usepackage{float}
\usepackage{array}
\usepackage{url}

\usepackage[final]{review} 
\setrevision{1}

\newcommand{\tof}{TOF}
\newcommand{\lidar}{LIDAR}
\newcommand{\spad}{SPAD}
\newcommand{\apd}{APD}

\begin{document}

\title{An Overview of Depth Cameras and Range Scanners Based on Time-of-Flight Technologies\thanks{This work has received funding from the French Agence Nationale de la Recherche (ANR) under the MIXCAM project ANR-13-BS02-0010-01, and from the European Research Council (ERC) under the Advanced Grant VHIA project 340113.}}

\titlerunning{An Overview of Depth Cameras and Range Scanners Based on Time-of-Flight Technologies}

\author{Radu Horaud \and Miles Hansard \and Georgios Evangelidis \and Cl\'ement M\'enier}
\authorrunning{R. Horaud, M. Hansard, G. Evangelidis \& C. M\'enier}

\institute{
R. Horaud and G. Evangelidis \at
INRIA Grenoble Rh\^one-Alpes\\
Montbonnot Saint-Martin,  France\\
\email{radu.horaud@inria.fr, georgios.evangelidis@inria.fr}
\and
M. Hansard \at
School of Electronic Engineering and Computer Science\\
Queen Mary University of London, United Kingdom\\
\email{miles.hansard@qmul.ac.uk}
\and
C. M\'enier \at
4D View Solutions\\
Grenoble, France\\
\email{clement.menier@4dviews.com}
}
\date{Received: date / Accepted: date}

\maketitle

\begin{abstract}
Time-of-flight (\tof) cameras are sensors that can measure the depths of scene-points, by illuminating the scene with a controlled laser or LED source, and then analyzing the reflected light. In this paper we will first describe the underlying measurement principles of time-of-flight cameras, including: (i)~pulsed-light cameras, which measure \textit{directly} the time taken for a light pulse to travel from the device to the object and back again, and (ii)~continuous-wave modulated-light cameras, which measure the phase difference between the emitted and received signals, and hence obtain the travel time \textit{indirectly}. We review the main existing designs, including prototypes as well as commercially available devices. We also review the relevant camera calibration principles, and how they are applied to TOF devices. Finally, we discuss the benefits and challenges of combined TOF and color camera systems.
\end{abstract}

\section{Introduction}
\label{section:introduction}
During the last decades, there has been a strong interest in the design and development of range-sensing systems and devices. 
The ability to remotely measure range is extremely useful and has been extensively used for mapping and surveying, on board of ground vehicles, aircraft, spacecraft and satellites, and for civil as well as military purposes. NASA has identified range sensing as a key technology for enabling autonomous and precise planet landing with both robotic and crewed space missions, e.g., \cite{amzajerdian2011lidar}. More recently, range sensors of various kinds have been used in computer graphics and computer vision for 3-D object modeling \cite{blais2004review}, Other applications
include terrain measurement, simultaneous localization and mapping (SLAM), autonomous
and semi-autonomous vehicle guidance (including obstacle detection), as well as object grasping and manipulation. Moreover, in computer vision, range sensors are ubiquitous in a number of applications, including object recognition, human motion capture, human-computer interaction, and 3-D reconstruction  \cite{Grzegorzek2013time}.

There are several physical principles and associated technologies that enable the fabrication of range sensors. One type of range sensor is known as \lidar{}, which stands either for \textit{``Light Imaging Detection And Ranging"} or for \textit{``LIght and ra\-DAR"}. \lidar{} is a remote-sensing technology that estimates range (or distance, or depth) by illuminating an object with a collimated laser beam, followed by detecting the reflected light using a photodetector. This remote measurement principle is also known as \textit{time of flight} (\tof{}).
Because \lidar{}s use a fine laser-beam, they can estimate distance with high resolution. \lidar{}s can use ultraviolet, visible or infrared light. They can target a wide range of materials, including metallic and non-metallic objects, rocks, vegetation, rain, clouds, and so on -- but excluding highly specular materials. 

The vast majority of range-sensing applications require an array of depth measurements, not just a single depth value. Therefore, \lidar{} technology must be combined with some form of scanning, such as a rotating mirror, in order to obtain a row of horizontally adjacent depth values. Vertical depth values can be obtained by using two single-axis rotating mirrors, by employing several laser beams with their dedicated light detectors, or by using mirrors at fixed orientations. In all cases, both the vertical field of view and the vertical resolution are inherently limited. Alternatively, it is possible to design a scannerless device: the light coming from a single emitter is diverged such that the entire scene of interest is illuminated, and the reflected light is imaged onto a two-dimensional array of photodetectors, namely a \tof{} \textit{depth camera}. Rather than measuring the intensity of the ambient light, as with standard cameras, \tof cameras measure the reflected light coming from the camera's own light-source emitter.

Therefore, both \tof{} range scanners and cameras belong to a more general category of \lidar{}s that combine a single or multiple laser beams, possibly mounted onto a rotating mechanism, with a 2D array of light detectors and time-to-digital converters, to produce 1-D or 2-D arrays of depth values.
Broadly speaking, there are two ways of measuring the time of flight \cite{remondino2013tof}, and hence two types of sensors:
\begin{itemize}
\item \textit{Pulsed-light} sensors directly measure the round-trip time of a light pulse. The width of the light pulse is of a few nanoseconds. Because the pulse irradiance power is much higher than the background (ambient) irradiance power, this type of \tof{} camera can perform outdoors, under adverse conditions, and can take long-distance measurements (from a few meters up to several kilometers). Light-pulse detectors are based on \textit{single photon avalanche diodes} (\spad{}) for their ability to capture individual photons with high time-of-arrival resolution \cite{cova1981towards}, approximatively 10~picoseconds ($10^{-11}$~s).
\item \textit{Continuous-wave} (CW) modulation sensors measure the phase differences between an emitted continuous sinusoidal  light-wave signal and the backscattered signals received by each photodetector \cite{lange2001solid}. 
The phase difference between emitted and received signals is estimated via cross-correlation (demodulation). The phase is directly related to distance, given the known modulation frequency. These sensors usually operate indoors, and are capable of short-distance measurements only (from a few centimeters to several meters). One major shortcoming of this type of depth camera is the phase-wrapping ambiguity \cite{hansard-2013}.
\end{itemize}

This paper overviews pulsed-light (section~\ref{section:pulsed-light}) and continuous wave (section~\ref{section:CW}) range technologies, their underlying physical principles, design, scanning mechanisms, advantages, and limitations. \addnote[commercial]{1}{We review the principal technical characteristics of some of the commercially available \tof{} scanners and cameras as well as of some laboratory prototypes.} Then we discuss the geometric and optical models together with the associated camera calibration techniques that allow to map raw \tof{} measurements onto Cartesian coordinates and hence to build 3D images or point clouds (section \ref{section:tof-calibration}). We also address the problem of how to combine \tof{} and color cameras for depth-color fusion and depth-stereo fusion (section~\ref{section:cross-calibration}). 

\section{Pulsed-Light Technology}
\label{section:pulsed-light}

As already mentioned, pulsed-light depth sensors are composed of both a light emitter and light receivers. The sensor sends out pulses of light emitted by a laser or by a laser-diode (LD). Once reflected onto an object, the light pulses are detected by an array of photodiodes that are combined with time-to-digital converters (TDCs) or with time-to-amplitude circuitry. There are two possible setups which will be referred to as \textit{range scanner} and \textit{3D flash \lidar{} camera}:
\begin{itemize}
\item 
A \textit{\tof{} range scanner} is composed of a single laser that fires onto a single-axis rotating mirror. This enables a very wide field of view in one direction (up to $360^\circ$) and a very narrow field of view in the other direction. One example of such a range scanner is the Velodyne family of sensors \cite{schwarz2010mapping} that feature a rotating head equipped with several (16, 32 or 64) LDs, each LD having its own dedicated photodetector, and each laser-detector pair being precisely aligned at a predetermined vertical angle, thus giving a wide vertical field of view. Another example of this type of \tof{} scanning device was recently developed by the Toyota Central R\&D Laboratories: the sensor is based on a single laser combined with a multi-facet polygonal (rotating) mirror. Each polygonal facet has a slightly different tilt angle, as a result each facet of the mirror reflects the laser beam into a different vertical direction, thus enhancing the vertical field-of-view resolution \cite{niclass2013100}.
\item
A \textit{3D flash \lidar{} camera} uses the light beam from a single laser that is \textit{spread} using an optical diffuser,  in order to illuminate the entire scene of interest. A 1-D or 2-D array of photo-detectors is then used to obtain a depth image. A number of sensor prototypes were developed using single photon avalanche diodes (\spad{}s) integrated with conventional CMOS timing circuitry, e.g., a 1-D array of 64 elements \cite{niclass2005design}, 32$\times$32 arrays \cite{albota2002three,stoppa2007cmos}, or a 128$\times$128 array \cite{niclass2008128}. A 3-D \textit{Flash \lidar{} camera}, featuring a 128$\times$128 array of \spad{}s, described in \cite{amzajerdian2011lidar}, is commercially available.
\end{itemize}

\subsection{Avalanche Photodiodes}
\label{section:spad}
One of the basic elements of any \tof{} camera is an array of photodetectors, each detector has its own timing circuit to measure the range to the corresponding point in the scene. Once a light pulse is emitted by a laser and reflected onto an object, only a fraction of the optical energy  is received by the detector -- \textit{the energy fall-off is inversely proportional to the square of the distance}. When an optical diffuser is present, this energy is further divided among multiple detectors. If a depth precision of a few centimeters is needed, the timing precision must be less than a nanosecond.\footnote{As the light travels at $ 3\times 10^{10}$ cm/s, 1 ns (or $10^{-9}$ s) corresponds to 30cm.} Moreover, the bandwidth of the detection circuit must be high, which also means that the noise is high, thus competing with the weak signal.

The vast majority of pulse-light receivers are based on arrays of \textit{single photon avalanche diodes} which are also referred to as \textit{Geiger-mode avalanche photodiodes} (G-\apd{}). A \spad{} is a special way of operating an avalanche photodiode  (\apd{}), namely it produces a fast electrical pulse of several volts amplitude in response to the detection of a single photon. This electrical pulse then generally trigers a digital CMOS circuit integrated into each pixel. An integrated \spad{}-CMOS array is a compact, low-power and all-solid-state sensor \cite{niclass2008128}.

The elementary building block of semiconductor diodes and hence of photodiodes is the 
$p-n$ junction, namely the boundary between two types of semiconductor materials, $p$-type and $n$-type. This is created by \textit{doping} which is a process that consists of adding impurities into an extremely pure semiconductor for the purpose of modifying its electrical (conductivity) properties. 
Materials conduct electricity if they contain mobile charge carriers. There are two types of charge carriers in a semiconductor: \textit{free electrons} and \textit{electron holes}. When an electric field exists in the vicinity of the junction, it keeps \textit{free electrons} confined on the $n$-side and \textit{electron holes} confined on the $p$-side. 

There are two types of $p-n$ diodes: forward-bias and reverse-bias. 
In forward-bias, the $p$-side is connected with the positive terminal of an electric power supply and the $n$-side is connected with the negative terminal. Both electrons and holes are pushed towards the junction. In reverse-bias, the $p$-side is connected with the negative terminal and the $n$-side is connected with the positive terminal. Otherwise said, the voltage on the $n$-side is higher than the voltage on the $p$-side. In this case both electrons and holes are pushed away from the junction. 

A photodiode is a semiconductor diode that converts light into current. The current is generated when photons are absorbed in the photodiode. Photodiodes are similar to regular semiconductor diodes. A photodiode is designed to operate in reverse bias. An \apd{} is a variation of a $p-n$ junction photodiode.
When an incident photon of sufficient energy is absorbed in the region where the field exists, an electron-hole pair is generated. Under the influence of the field, the electron drifts to the $n$-side and the hole drifts to the $p$-side, resulting in the flow of \textit{photocurrent} (i.e., the current induced by the detection of photons) in the external circuit. When a photodiode is used to detect light, the number of electron-hole pairs generated per incident photon is at best unity.

An \apd{} detects light by using the same principle \cite{aull2002geiger}.
The difference between an \apd{} and an ordinary $p-n$
junction photodiode is that an \apd{} is designed to
support high electric fields. When an electron-hole
pair is generated by photon absorption, the electron
(or the hole) can accelerate and gain sufficient energy
from the field to collide with the crystal lattice and
generate another electron-hole pair, losing some of its
kinetic energy in the process. This process is known as
\textit{impact ionization}. The electron can accelerate again,
as can the secondary electron or hole, and create more
electron-hole pairs, hence the term ``avalanche".

\begin{figure}
\begin{center}
\includegraphics[width=0.98\linewidth]{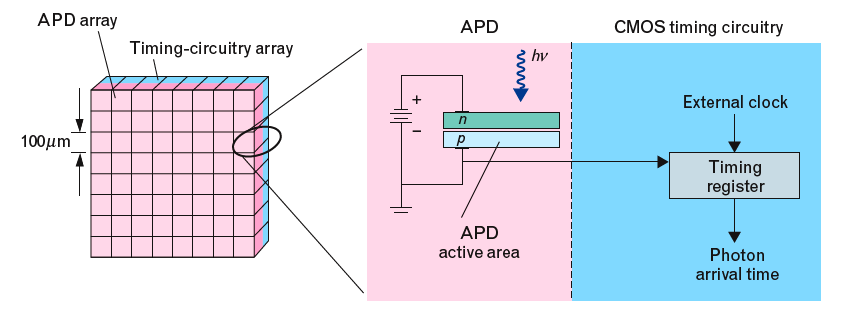} 
\caption{\label{fig:spad} 
The basic structure of an array of time of flight detectors consists of Geiger-mode APDs (pink) bonded to complementary-metal-oxide-semiconductor (CMOS) timing circuitry (blue). A photon of energy \textit{hv} is absorbed in the APD
active area. The gain of the APD resulting from the electron avalanche is great enough that the detector
generates a pulse that can directly trigger the 3.3 V CMOS circuitry. No analog-to-digital converter is
needed. A digital logic latch is used to stop a digital timing register for each pixel. The time of flight is recorded
in the digital value of the timing register.}
\end{center}
\end{figure}

After
a few transit times, a competition develops between
the rate at which electron-hole pairs are being generated
by \textit{impact ionization} (analogous to a birth rate)
and the rate at which they exit the high-field region
and are \textit{collected} (analogous to a death rate). If the
magnitude of the reverse-bias voltage is below a value
known as the \textit{breakdown voltage}, collection wins the
competition, causing the population of electrons and
holes to decline. If the magnitude of the voltage is above the breakdown voltage, impact ionization wins. This situation represents the most commonly
known mode of operation of \apd{}s: measuring the
intensity of an optical signal and taking advantage of
the internal gain provided by impact ionization. Each
absorbed photon creates on average a finite number
$M$ of electron-hole pairs. The internal gain $M$ is typically
tens or hundreds. Because the average photocurrent
is strictly proportional to the incident optical
flux, this mode of operation is known as linear mode.

\subsection{Single Photon Avalanche Diodes}
\label{subsection:SPAD}

The fundamental difference between \spad{}s (also referred to as Geiger-mode \apd{}) and conventional \apd{}s is that \spad{}s are specifically designed to operate with a reverse-bias voltage well above the breakdown voltage. A \spad{} is able to detect low intensity incoming light (down to the single photon) and to signal the arrival times of the photons with a jitter of a few tens of picoseconds \cite{cova1981towards}. Moreover, \spad{}s behave almost like digital devices, hence subsequent signal processing can be greatly simplified.
The basic structure of an array of time-of-flight detectors consists of \spad{}s bonded to CMOS timing circuitry. A \spad{} outputs an analog voltage pulse, that reflects the detection of a single photon, and that can directly trigger the CMOS circuitry.  The latter implements time-to-digital converters to compute time-interval measurements between a start signal, global to all the pixels, and photon arrival times in individual pixels. 

\section{\lidar{} Cameras}
\label{section:lidar-cameras}
\subsection{Velodyne Range Scanners}
\label{section:velodyne}
Whereas most LIDAR systems have a
single laser that fires onto a rotating mirror and hence are only able to view objects in a single plane,
the high-definition HDL-64E \lidar{} range scanner from
Velodyne\footnote{\url{http://velodynelidar.com/index.html}}
uses a rotating head featuring
64 semiconductor lasers. Each laser operates at 905 nm wavelength, has a beam divergence of 2~mrad, and fires 5~ns light pulses at up to $20,000$~Hz. The 64 lasers are spread over a vertical field of view,  and coupled with 64 dedicated photo detectors for precise ranging. The laser-detector pairs are precisely aligned at vertical angles to give a $26.8^{\circ}$ vertical field of view. By
spinning the entire unit at speeds of up to
900 rpm (15 Hz) around its vertical axis,
a 360$^\circ$ field-of-view is generated. 

\begin{figure}[h!]
 \includegraphics[width=0.25\textwidth]{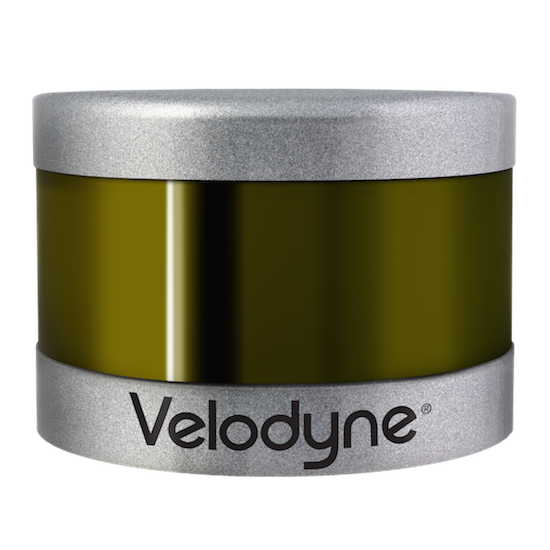} 
   \includegraphics[width=0.21\textwidth]{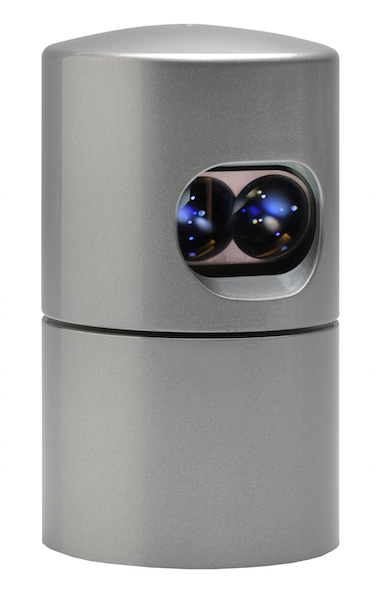} 
   \includegraphics[width=0.3\textwidth]{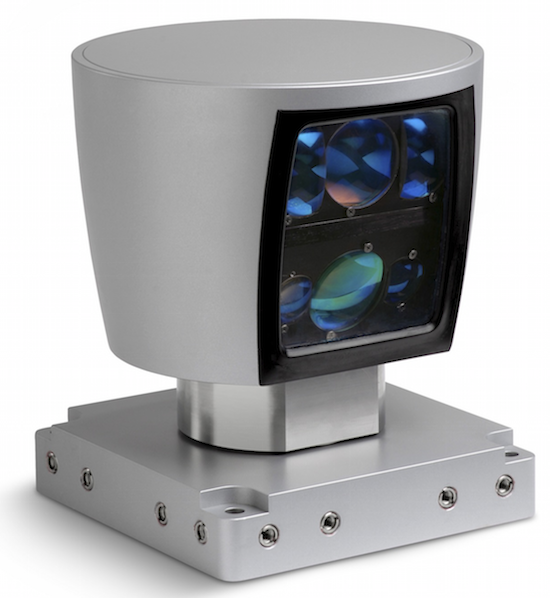}
\caption{\label{fig:velodyne} 
From left to right: VLP-16, HDL-32E, and HDL-64E high-definition range scanners manufactured by Velodyne. These rotating scanners feature a complete  360$^\circ$ horizontal field of view as well as 16, 32 and 64 laser-detector pairs spread over a vertical field of view. The sensors range is from 1 m and up to 120 m (depending on the material properties of the scanned object) with a accuracy of 2 cm.}
\end{figure}

This
allows the sensor to achieve data collection
rates that are an order of magnitude higher
than most conventional designs.
Over
1.3 million data points are generated each
second, independent of the spin rate. 
Velodyne also commercializes 32 and 16 laser scanners with reduced vertical resolution. The range scanners are shown on figure~\ref{fig:velodyne} and their main parameters and specifications are summarized on Table~\ref{table:velodyne}. 

A mathematical model of the Velodyne HDL-64E range scanner was developed in \cite{glennie2007rigorous,glennie2010static} together with a calibration model and a practical method for estimating the model parameters. Extensive experiments using this model show that the actual noise level of the range measurements is 3.0 to 3.5~cm, which is double the manufacturer specification. Subsequently, the same authors analyzed the temporal stability (horizontal angle offset) of the scanner \cite{glennie2011temporal}.

\subsection{Toyota's Hybrid \lidar{} Camera}
\label{section:toyota}

\begin{figure}
\begin{center}
   \includegraphics[height=0.38\linewidth]{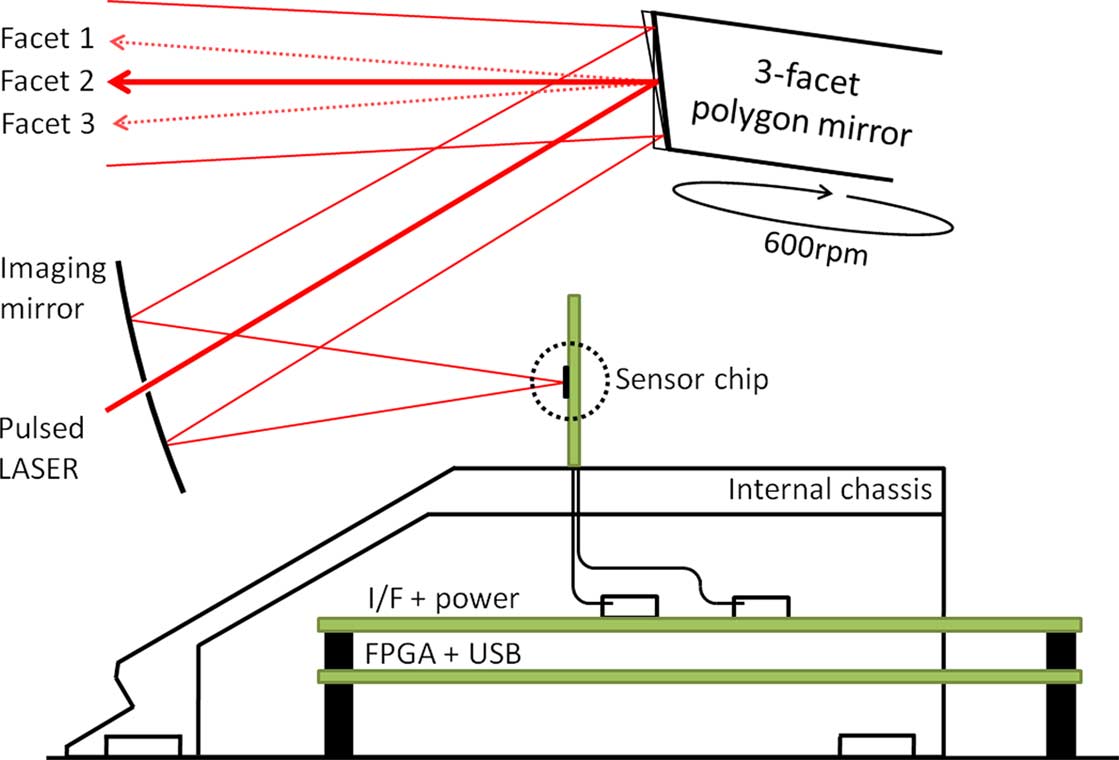} 
   \includegraphics[height=0.38\linewidth]{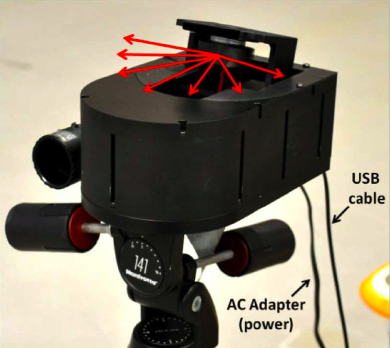}
\caption{\label{fig:toyota} A simplified diagram of a depth sensor system developed by Toyota (left) and a view of the complete system (right).}
\end{center}
\end{figure}

Fig.~\ref{fig:toyota} (left) shows a simplified diagram of the depth sensor system recently developed at the Toyota Central R\&D Labs, Japan \cite{niclass2013100}.
A 870~nm pulsed laser source with a repetition rate of 200~kHz emits an optical beam with
1.5$^\circ$ and 0.05$^\circ$ of divergence in the vertical and horizontal directions,
respectively. While the optical pulse duration is 4~ns
full-width at half-maximum (FWHM), the mean optical power
is 40~mW. The laser beam is coaxially aimed at the three-facet
polygonal mirror through an opening in the center of an imaging
concave mirror. Each facet of the polygonal mirror has a slightly
different tilt angle. 
\begin{table*}
\begin{center}
\begin{tabular}{|l|l|l|l|l|r|l|l|}
\hline 
Model & Resolution (H$\times$V) & Range/Accuracy & FOV & Frame rate & Points/second & Laser & Pulse width \\
\hline \hline
HDL-64E & $0.08^\circ\times 0.4^\circ$ & 2 -- 120 m / 2 cm & $360^\circ\times 26.8^\circ$ & 5-15 Hz & 1,300,000 & 905 nm & 10 ns \\
\hline
HDL-32E & $0.08^\circ\times 1.33^\circ$ & 2 -- 120 m / 2 cm & $360^\circ\times 31.4^\circ$ & 5-20 Hz & 700,000 & 905 nm & 10 ns \\
\hline
VLP-16 & $0.08^\circ\times 1.87^\circ$ & 2 -- 100 m / 2 cm & $360^\circ\times 30^\circ$ & 5-20 Hz & 300,000 & 905 nm & 10 ns \\
\hline
Toyota & $0.05^\circ\times 1.5^\circ$ & not specified & $170^\circ\times 4.5^\circ$ & 10 Hz & 326,400 & 870 nm & 4 ns \\
\hline
\end{tabular}
\end{center}
\caption{\label{table:velodyne} The principal characteristics of the Velodyne \lidar{} range scanners and of Toyota's \lidar{} prototype that can operate outdoors. The maximum range depends on the material properties of the targeted object and can vary from 50~m (for pavement) to 120~m (for cars and trees). All these range scanners use class 1 (eye safe) semiconductor lasers.}
\end{table*}

As a result, in one revolution of 100~ms,
the polygonal mirror reflects the laser beam into three vertical
directions at $+1.5^\circ$, 0$^\circ$, and $-1.5^\circ$, thus covering, together with
the laser vertical divergence, a contiguous vertical FOV of 4.5$^\circ$.
During the 170$^\circ$ horizontal scanning, at one particular facet,
back-reflected photons from the targets in the scene are collected
by the same facet and imaged onto the CMOS sensor
chip at the focal plane of the concave mirror. The chip has a
vertical line sensor with 32 macro-pixels. These pixels resolve
different vertical portions of the scene at different facet times,
thus generating an actual vertical resolution of 96 pixels. Since
each macro-pixel circuit operates in full parallelism, at the end
of a complete revolution, 1020$\times$32 distance points are computed.
This image frame is then repartitioned into 340$\times$96 actual
pixels at 10~FPS. An optical near-infrared interference
filter (not shown in the figure) is also
placed in front of the sensor for background light rejection.

The system electronics consists of a rigid-flex head-sensor
PCB, a laser driver board, a board for signal interface and
power supply, and a digital board comprising a low-cost FPGA
and USB transceiver. Distance, intensity, and reliability data are
generated on the FPGA and transferred to a PC at a moderate
data rate of 10 Mbit/s. The system requires only a compact
external AC adapter from which several other power supplies
are derived internally.

\subsection{3D Flash \lidar{} Cameras}
\label{section:flash}
\begin{figure}[t!]
\begin{center}
\includegraphics[width=0.95\linewidth]{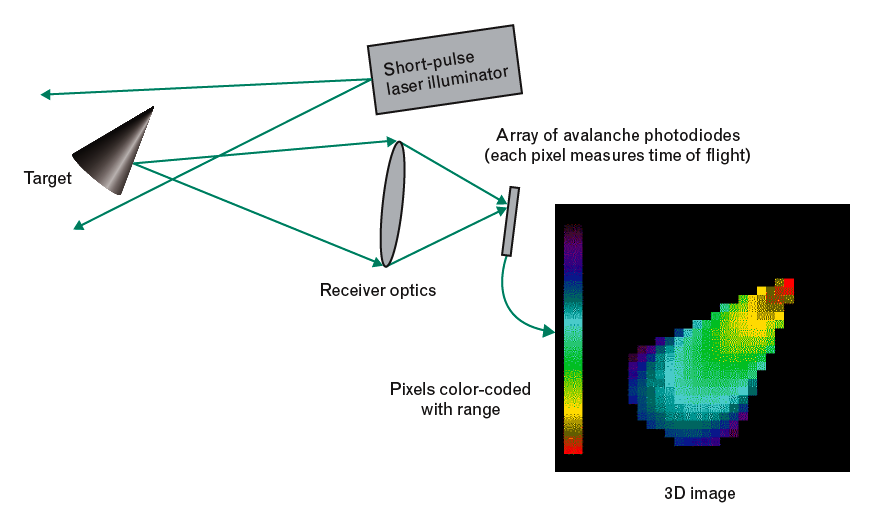} 
\caption{\label{fig:flashlidar} This figure illustrates the basic principle of a Flash \lidar{} camera. An eye-safe laser flood-illuminates an object of interest.}
\end{center}
\end{figure}

A 3D Flash \lidar{} is another name used  to designate a sensor that creates a 3D image (a depth value at each pixel) from a single laser pulse that is used to \textit{flood-illuminate} the targeted scene or objects. The main difference between a \lidar{} camera and a standard \lidar{} device is that there is no need of a mechanical scanning mechanism, e.g., rotating mirror. 
Hence, a Flash \lidar{} may well be viewed as a 3D \textit{video} camera that delivers 3D images at up to 30 FPS. 
The general principle and basic components are shown of Fig.~\ref{fig:flashlidar}.
Flash \lidar{}s use a light-pulse emitted by a single laser, that is reflected  onto a scene object. Because the reflected light is further divided among multiple detectors, the energy fall-off is considerable. Nevertheless, the fact that there is no need for scanning represents a considerable advantage. Indeed, each individual \spad{} is exposed to the optical signal for a long period of time, typically of the order of ten milliseconds. This allows for a large number of illumination cycles that can be averaged to reduce the various effects of noise.

The MIT Lincoln Laboratory reported the development of 3D Flash \lidar{} long-range (up to 500 m) camera prototypes based on a short-pulse (1 ns) microchip laser, transmitting at a wavelength of 532 nm, and \spad{}/CMOS imagers  \cite{albota2002three,aull2002geiger}. Two \lidar{} camera prototypes were developed at MIT Lincoln Laboratory, one based on a 4$\times$4 pixels \spad{}/CMOS sensor combined with a two-axis rotating mirror, and one based on a 32$\times$32 pixels \spad{}/CMOS sensor.

\begin{figure}
\begin{center}
\begin{tabular}{cc}
  \includegraphics[width=0.46\linewidth]{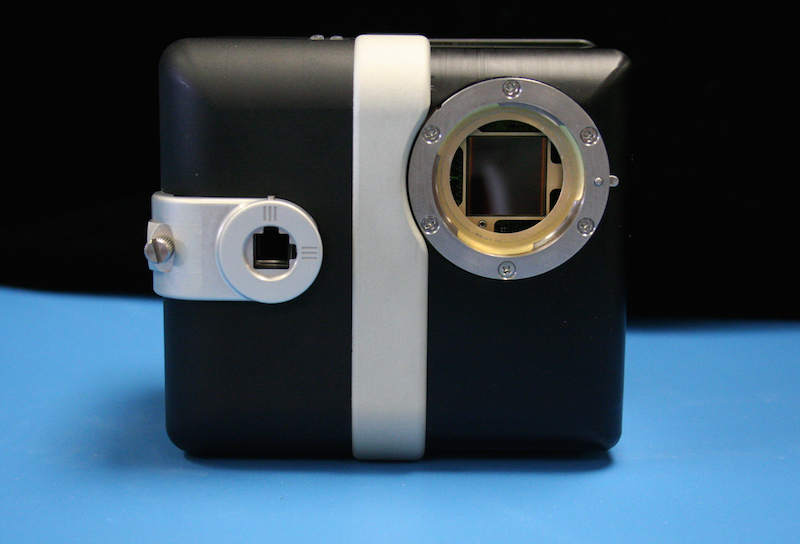} &
   \includegraphics[width=0.46\linewidth]{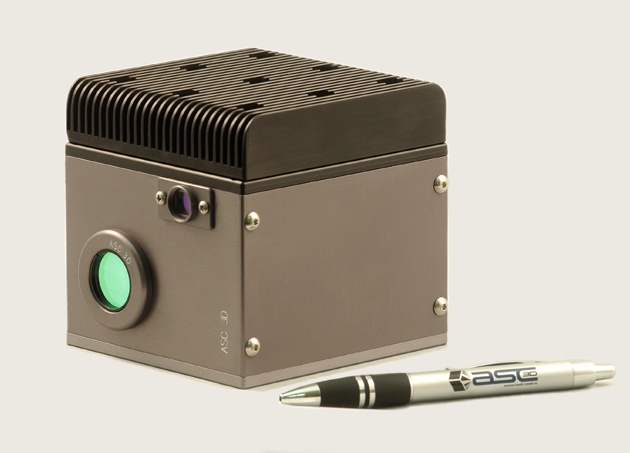} \\
   Tiger Eye (new) & Tiger Eye (old) \\
      \includegraphics[width=0.46\linewidth]{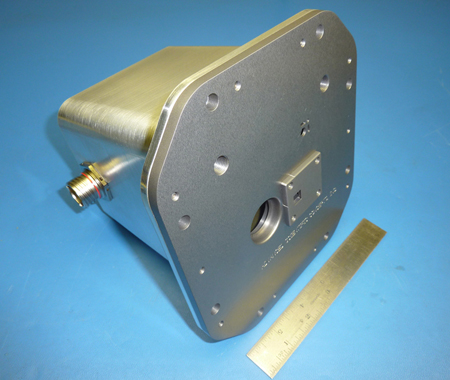} &
      \includegraphics[width=0.46\linewidth]{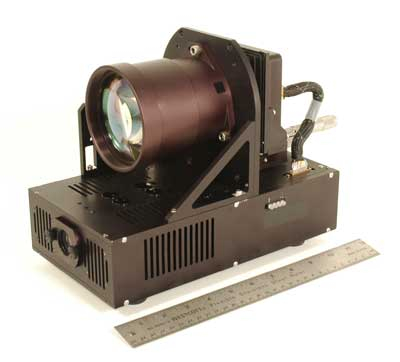} \\
  Dragon Eye & Portable 3D 
\end{tabular}
\end{center}
\caption{\label{fig:dragoneye} 3D Flash \lidar{} depth cameras manufactured and commercialized by Advanced Scientific Concepts, Santa Barbara, CA. These cameras use a single pulsed-light diffused laser beam and can operate outdoors in adverse conditions at up to 30FPS.}
\end{figure}

Advanced Scientific Concepts Inc.\footnote{\url{http://www.advancedscientificconcepts.com/index.html}} developed a 3D Flash \lidar{} prototype \cite{stettner2008three} as well as a number of commercially available \lidar{} cameras, e.g., Fig.~\ref{fig:dragoneye}. The TigerEye/TigerCup\footnote{\url{http://www.advancedscientificconcepts.com/products/tigercub.html}} is truly a 3D video camera. It is equipped with an eye-safe 1570 nm laser, with a CMOS 128$\times$128 pixels sensor, and it delivers images at 10-30 FPS. It has an interchangeable lens such that its range and field of view (FOV) can vary: 3$^\circ \times 3^\circ$ FOV and range up to 1100 meters, 
8.6$^\circ \times 8.6^\circ$ FOV and
range up to 450 meters, 45$^\circ\times 22^\circ$ FOV and
range up to 150 meters, and 45$^\circ \times 45^\circ$ FOV and
range up to 60 meters.

The DragonEye/GoldenEye\footnote{\url{http://www.advancedscientificconcepts.com/products/portable.html}} 3D Flash \lidar{} space camera delivers both intensity and depth videos at 10~FPS. It has a 128$\times$128 \spad{} based sensor  and its FOV is of 45$^\circ\times 45^\circ$ which is equivalent of a 17 mm focal length and it can range up to 1500 m. The DragonEye was tested and used by NASA for precision navigation and safe landing \cite{amzajerdian2011lidar}. The specifications of these cameras are summarized in Table~\ref{3dflash-summary}.

\begin{table*}[t!]
\centering
\begin{tabular}{|l|l|l|l|l|l|l|l|}
\hline
Camera & Resolution & Range & Mult. cameras & FOV & FPS & Laser & Indoor/out\\
\hline \hline
TigerEye-1  & 128$\times$128 & 1100 m & not specified &$3^\circ\times 3^\circ$ &10-30 &1570 nm & no/yes \\ \hline
TigerEye-2  & 128$\times$128 & 450 m & not specified &$ 8.6^\circ\times 8.6^\circ$ &10-30 &1570 nm & no/yes \\ \hline
TigerEye-3  & 128$\times$128 & 150 m & not specified &$45^\circ\times 22^\circ$ &10-30 &1570 nm & no/yes \\ \hline
TigerEye-4  & 128$\times$128 & 60 m & not specified &$45^\circ\times 45^\circ$ &10-30 &1570 nm & no/yes \\ \hline
DragonEye  & 128$\times$128 & 1500 m & not specified &$45^\circ\times 45^\circ$ &10-30 &1570 nm & no/yes \\ \hline
Real.iZ VS-1000 & 1280$\times$1024 & 10 m & possible &$45^\circ\times 45^\circ$ & 30-450 & 905 nm & yes/yes \\ \hline
Basler & 640$\times$480 & 6.6 m & not specified &$57^\circ\times 43^\circ$ & 15 & not specified & yes/yes \\ \hline
\end{tabular}
\caption{\label{3dflash-summary} This table summarizes the main features of commercially available 3D Flash \lidar{} cameras. The accuracy of the depth measurements announced by the manufacturers are not reported in this table as the precision of the measurements depend on a lot of factors, such as the surface properties of the scene objects, illumination conditions, frame rate, etc.}
\end{table*}

\subsection{Other \lidar{} camera}
\label{section:odos}

\begin{figure}
\centering
   \includegraphics[width=0.6\linewidth]{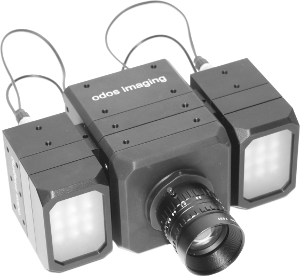} 
   \caption{\label{fig:odos} The high-resolution  real.iZ pulsed-light \lidar{} camera manufactured by Odos Imaging.}
   \end{figure}

Recently, Odos Imaging\footnote{\url{http://www.odos-imaging.com/}} announced the commercialization of a high-resolution pulsed-light time-of-flight camera, Fig.~\ref{fig:odos}. 
The camera has a resolution of 1280$\times$1024 pixels, a range up to 10 m, a frame rate of 30~FPS  and up to 450~FPS, depending on the required precision (Table~\ref{3dflash-summary}. It can be used both indoor and outdoor (for outdoor applications it may require additional filters). One advantage of this camera is that it delivers both depth and standard monochrome images. Another \lidar{} camera is Basler's pulsed-light camera based on a Panasonic TOF CCD sensor. The main characteristics of these cameras are summarized in Table~\ref{3dflash-summary}.

\section{Continuous-Wave Technology}
\label{section:CW}
\begin{figure}[t!]
\centering
   \includegraphics[width=0.97\linewidth]{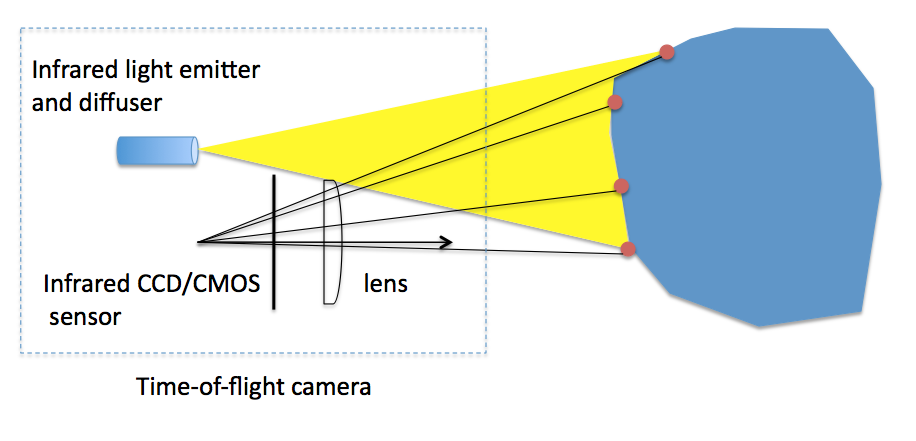} 
   \caption{\label{fig:cw-tof} This figure shows the image formation principle of 3D Flash \lidar{}s and continuous-wave TOF cameras. }
   \end{figure}

All these depth sensors share some common characteristics, as follows \cite{lange2001solid,remondino2013tof}:
\begin{itemize}
\item The transmitter, a light emitter (generally a LED, or light-emitting diode) sends light onto an object and the time the light needs to travel from the illumination source to the object and back to the sensor is measured. 
\item In the case of \textit{continuous-wave} (CW), the emitted signal is a sinusoidally modulated light signal. 
\item The received signal is \textit{phase-shifted} due to the round trip of the light signal. Moreover, the received signal is affected by the object's reflectivity, attenuation along the optical path and background illumination.
\item Each pixel independently performs demodulation of the received signal and therefore is capable of measuring both its phase delay as well as amplitude and offset (background illumination). 
\end{itemize}
The imaging principle of a CW-TOF camera is shown on Fig.~\ref{fig:cw-tof}. 

\subsection{Demodulation Principle}
\label{sec:demodulation}

\begin{figure*}
\centering
   \includegraphics[width=0.8\linewidth]{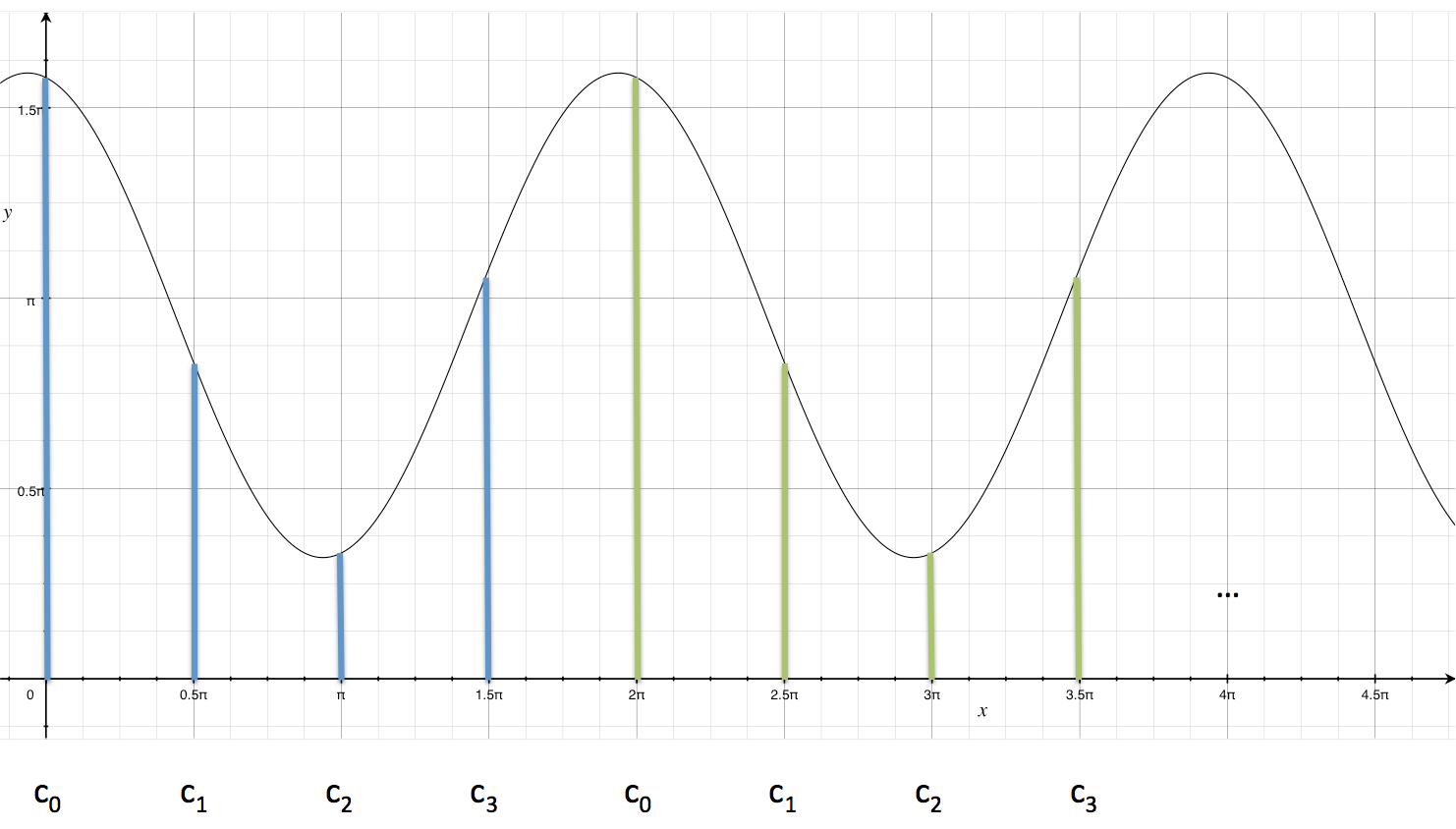} 
   \caption{\label{fig:4bucket} This figure shows the general principle of the four-bucket method that estimates the demodulated optical signal at four equally-spaced samples in one modulation period. A CCD/CMOS circuit achieves light detection, demodulation, and charge storage. The demodulated signal is stored at four equally spaced samples in one modulation period. From these four values, it is then possible to estimate the phase and amplitude of the received signal as well as the amount of background light (offset).}
   \end{figure*}

Let $s(t)$ and $r(t)$ be the optical powers of the emitted and received signals respectively:
\begin{align}
\label{eq:emitted-signal}
s(t) &= a_1 + a_2 \cos (2 \pi f t),\\
\label{eq:received-signal}
r(t) &= A \cos (2\pi f t - 2 \pi f \tau) + B,
\end{align}
where $f$ is the modulation frequency, $\tau$ is the time delay between the emitted and received signals, $\phi = 2\pi f \tau$ is the corresponding phase shift, $a_1$ and $a_2$ are the offset and amplitude of the modulated emitted signal, $A$ is the amplitude of the received signal, and $B$ is the offset of the received signal due to background illumination (illumination other than the emitter itself). The cross-correlation between the powers of the emitted and received signals can be written as:
\begin{equation}
\mathcal{C}(x) = \lim_{T\rightarrow \infty} \frac{1}{T} \int_{-T/2}^{+T/2} s(t) r(t-x) \mathrm{d}t.
\end{equation}
By substituting $s(t)$ and $r(t)$ with their expressions \eqref{eq:emitted-signal} and \eqref{eq:received-signal} and by developing the terms, we obtain:
\begin{align}
\mathcal{C}(x,\tau) = & \lim_{T\rightarrow \infty} \frac{1}{T} \int_{-T/2}^{+T/2} \Big(  a_2 B \cos (2\pi f t) \nonumber \\
&+ a_2 A \cos (2\pi f t) \cos (2\pi f t - 2\pi f (\tau + x)) \nonumber \\
&+ a_1 A \cos (2\pi f t - 2\pi f (\tau + x)) \Big) \mathrm{d}t  \nonumber \\
&+ a_1 B.
\end{align}
Using the identities 
\begin{align*}
\lim_{T\rightarrow \infty} \frac{1}{T} \int_{-T/2}^{+T/2} \cos t \; \mathrm{d}t & =0 \\
\lim_{T\rightarrow \infty} \frac{1}{T} \int_{-T/2}^{+T/2} \cos t \cos (t-u) \; \mathrm{d}t & = \frac{1}{2} \cos u,
\end{align*}
we obtain:
\begin{align}
\mathcal{C}(x,\tau) &= \frac{a_2 A}{2} \cos ( 2\pi f (x + \tau) ) + a_1B.
\end{align}
Using the notations $\psi=2\pi f x$ and $\phi = 2\pi f  \tau$, we can write:
\begin{equation}
\mathcal{C}(\psi, \phi) = \frac{a_2 A}{2} \cos ( \psi + \phi ) + a_1 B.
\end{equation}
Let's consider the values of the correlation function at four equally spaced samples within one modulation period, $\psi_0 = 0, \psi_1 = \pi / 2, \psi_2 = \pi$, and $\psi_3 = 3\pi / 2$, namely $C_0=\mathcal{C}(0,\phi),C_1=\mathcal{C}(\pi/2,\phi),C_2=\mathcal{C}(\pi,\phi)$, and $C_3=\mathcal{C}(3\pi/2,\phi)$, e.g., Fig.~\ref{fig:4bucket}. These four sample values are sufficient for the unambiguous computation of the offset $B$, amplitude $A$, and phase $\phi $ \cite{lange2001solid}:
\begin{align}
\phi &= \arctan \left(  \frac{C_3- C_1}{C_0- C_2} \right)\\
A &= \frac{1}{a_2} \sqrt {(C_3- C_1)^2 + (C_0- C_2)^2}\\
B &=  \frac{1}{4a_1} \left( C_0 +  C_1 + C_2 +  C_3 \right)
\end{align}

\subsection{Pixel Structure}
An electro-optical demodulation pixel performs the following operations, \cite{buttgen2008robust}: 
\begin{itemize}
\item light detection, the incoming photons are converted into electron charges;
\item demodulation (based on correlation),
\item clocking, and
\item charge storage. 
\end{itemize}

The output voltage of the storage capacitor, after integration over a short period of time $T_i$, is proportional to the correlation ($R$ is the optical responsivity of the detector):
\begin{equation}
V(x) = \frac{R T_i}{C_S} C(x)
\end{equation}
Hence, the received optical signal is converted into a photocurrent. 

The samples are the result of the integration of the photocurrent of a duration $\Delta t< 1/f$. In order to increase the signal-to-noise ratio of one sampling process, the samples $C_0$ to $C_3$ are the result of the summation over many modulation periods (up to hundreds of thousands). 
A pixel with four shutters feeding four charge storage nodes allows the simultaneous acquisition of the four samples needed for these computations. The four shutters are activated one at a time for a time equal to $T/4$ (where $T$ is the modulation period), and the shutter activation sequence is repeated for the whole integration time $T_i$ which usually includes hundreds of thousands of modulation periods.

\subsection{Depth Estimation from Phase}
A depth value $d$ \textbf{at each pixel} is computed with the following formula:
\begin{equation}
d = \frac{1}{2} c\tau 
\end{equation}
where $c$ is the light speed and $\tau$ is the time of flight. Since we measure the phase $\phi = 2\pi f \tau$, we obtain:
\begin{equation}
d =  \frac{c }{4 \pi f} \phi = \frac{c }{4 \pi f} \arctan \left(  \frac{C_3- C_1}{C_0- C_1} \right)
\end{equation}
Nevertheless, because the phase is defined up to $2\pi$, there is an inherent \textbf{phase wrapping ambiguity} in measuring the depth:
\begin{itemize}
\item Minimum depth: $d_{\min} = 0 \;\; (\phi = 0)$
\item Maximum depth: $d_{\max} = \frac{c}{2f} \;\; (\phi = 2\pi)$.
\end{itemize}
The depth at each pixel location $(i,j)$ can be written as a function of this wrapping ambiguity:
\begin{equation}
d(i,j) = \left( \frac{\phi(i,j)}{2\pi} + n(i,j)\right) d_{\max} 
\end{equation}
where $n=0,1,2,\hdots$ is the number of wrappings. This can also be written as:
\begin{equation}
d(i,j) = d_{\tof}(i,j) + n(i,j) d_{\max}
\end{equation}
where $d$ is the \textit{real} depth value and $d_{\tof}$ is the \textit{measured} depth value. It is important to stress that the number of wrappings is not the same at each pixel. Let's consider a modulation frequency $f=30$~MHz, the unambiguous range of the camera is in this case from $d_{\min}=0$ to $d_{\max}=5$ meters. The ambiguity decreases as the modulation frequency decreases, but in the same time the accuracy decreases as well. Several methods were proposed in the literature to solve for the phase wrapping ambiguity \cite{Ghiglia94,Opri07,Bioucas-Dias07,Payne09,McClure10,Droeschel10b,Droeschel10a,Choi10,Choi12a}

To summarize, the following remarks can be made concerning these depth-camera technologies:
\begin{itemize}
\item A CW-TOF camera works at a very precise modulation frequency. Consequently, it is possible to simultaneously and synchronously use several CW-TOF cameras, either by using a different modulation frequency for each one of the cameras, e.g., six cameras in the case of the SR4000 (Swiss Ranger), or by encoding the modulation frequency, e.g., an arbitrary number of SR4500 cameras.
\item In order to increase the signal-to-noise ratio, and hence the depth accuracy, CW-TOF cameras need a relatively long integration time (IT), over several time periods. In turn, this introduces \textit{motion blur} \cite{hansard-2013} (chapter~1) in the presence of moving objects. Because of the need of long IT, fast shutter speeds (as done with standard cameras) cannot be envisaged. 
\end{itemize}
To summarize, the sources of errors of these cameras are: demodulation, integration, temperature, motion blur, distance to the target, background illumination, phase wrapping ambiguity, light scattering, and multiple path effects. A quantitative analysis of these sources of errors is available in \cite{foix-2011}. In the case of several cameras operating simultaneously, interferences between the different units is an important issue.

\section{\tof{} Cameras}
\label{section:tof-cameras}

\addnote[commercial-cw]{1}{In this section we review the characteristics of some of the commercially available cameras. We selected those camera models for which technical and scientific documentation is readily available. The main specifications of the overviewed camera models are summarized in Table~\ref{CW-summary}.
}
\subsection{The SR4000/SR4500 Cameras}
\label{sec:SR4000}
\begin{figure}
\centering
\begin{tabular}{cc}
\includegraphics[width=0.44\linewidth]{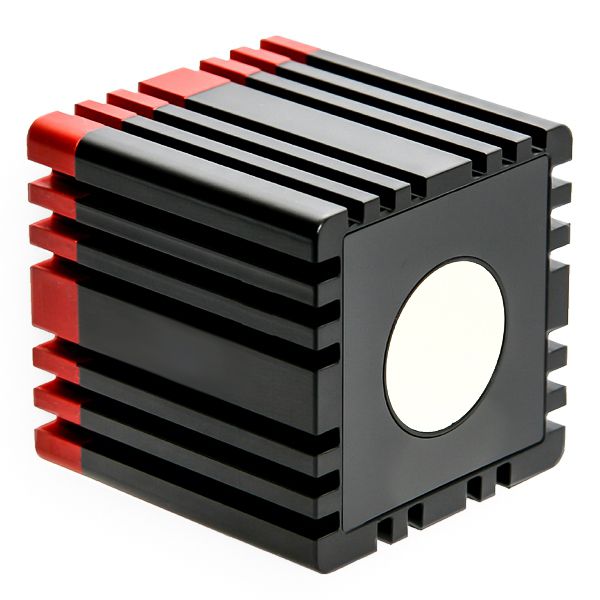} &
\includegraphics[width=0.44\linewidth]{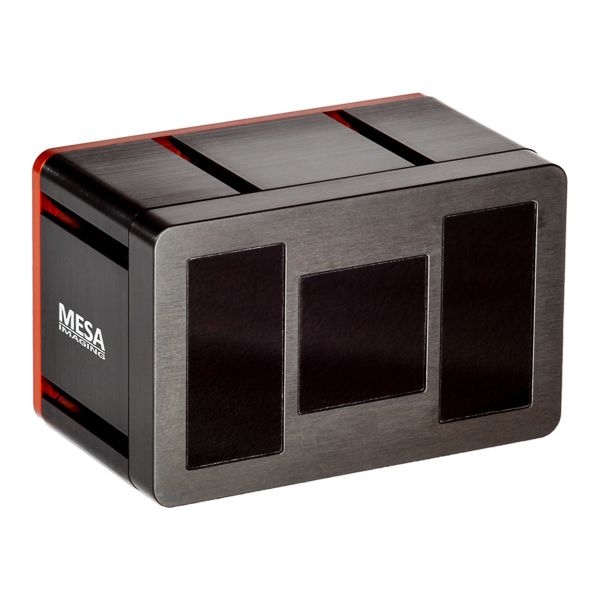} 
\end{tabular}
\caption{\label{fig:SR4000} The SR4000 (left) and SR4500 (right) CW-TOF cameras manufactured by Mesa Imaging.}
\end{figure}

The SR4000/4500 cameras, figure~\ref{fig:SR4000}, are manufactured by Mesa Imaging, Zurich, Switzerland.\footnote{\url{http://www.mesa-imaging.ch/}} They are con\-ti\-nuous-wave TOF cameras that provide depth, amplitude, and confidence images with a resolution of 176$\times$144 pixels. In principle, the cameras can work at up to 30 FPS but in practice more accurate depth measurements are obtained at 10-15~FPS. 

The modulation frequency of the SR4000 camera can be selected by the user. The camera can be operated at:
\begin{itemize}
\item $29$ MHz, 30 MHz, or $31$ MHz corresponding to a maximum depth of $5.17$ m, $5$ m and $4.84$ m respectively.
\item $14.5$ MHz, 15.0 MHz, or $15.5$ MHz corresponding to a maximum depth of $10.34$ m, $10$ m and $9.67$ m respectively.
\end{itemize}
This allows the simultaneous and synchronous use of up to six SR4000 cameras to be used together with any number of color cameras. 

The modulation frequency of SR4500 is of $13.5$ MHz which allows a maximum depth of 9~m. Moreover, an arbitrary number of SR4500 cameras can be combined together because the modulation frequency is encoded differently for each unit.

\begin{figure}
\centering
\begin{tabular}{cccc}
\includegraphics[width=0.44\linewidth]{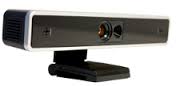} &
\includegraphics[width=0.44\linewidth]{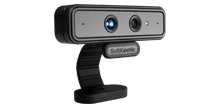}
\end{tabular}
\caption{\label{fig:DS311} The DS311 (left) and DS325 (rigtht) CW-TOF cameras manufactured by SoftKinetic.}
\end{figure}

\subsection{The Kinect v2 RGB-D Camera}

The Kinect color and depth (RGB-D) camera, manufactured by Microsoft, was recently upgraded  to Kinect~v2. Unlike the former version that was based on structured-light technology, the latter uses a time-of-flight sensor \cite{payne2014,bamji20150} and was mainly designed for gaming \cite{sell2014xbox}. \addnote[kinect-v2]{1}{Kinect-v2 achieves one of the best image resolution among TOF cameras commercially available. Moreover, it uses multiple modulation frequencies (10-130~MHz) thus achieving an excellent compromise between depth accuracy and phase unwrapping, i.e. Section 4.3 above. In \cite{bamji20150} it is reported that the Kinect~v2 can measure depth in the range 0.8-4.2~m with an accuracy of 0.5\% of the measured range. 
}
Several recent articles evaluate the Kinect v2 sensor for mobile robotics \cite{fankhauser2015kinect} and in comparison with the structured-light version \cite{sarbolandi-2015}.
 
It is interesting to note that Kinect v2 is heavier than its predecessor (970 g instead of 170 g) requires higher voltage (12 V instead of 5 V) and power usage (15 W instead of 2.5 W).
\subsection{Other CW TOF cameras}

The following depth cameras are based on the same continuous wave demodulation principles (see Table~\ref{CW-summary} for a summary of the characteristics of these cameras):
\begin{itemize}
\item DS311 and DS325 cameras, figure~\ref{fig:DS311}, manufactured by SoftKinetic,\footnote{\url{http://www.softkinetic.com/}}
\item E70 and E40 manufactured by Fotonic,\footnote{\url{http://www.fotonic.com/}}
\item TOF sensor chip manufactured by PMD.\footnote{\url{http://www.pmdtec.com/}},
\item The D-imager manufactured by Panasonic has a range up to 15 cm. It was discontinued in March 2015.\footnote{\url{http://www2.panasonic.biz/es/densetsu/device/3DImageSensor/en/}}
\end{itemize}

\begin{table*}[t!]
\centering
\begin{tabular}{|l|l|l|l|l|l|l|l|}
\hline
Camera & Resolution & Range & Mult. cameras & FOV & Max FPS & Illumination & Indoor/out\\
\hline \hline
SR4000  & 176$\times$144 & 0$-$5 or 0$-$10 m & 6 cameras &$43^\circ\times 34^\circ$ &30 & LED& yes/no \\ \hline
SR4500  & 176$\times$144  &0$-$9 m & many cameras & $43^\circ\times 34^\circ$&30 &LED & yes/no \\ \hline
DS311  & 160$\times$120  &0.15$-$1 or 1.5$-$4.5 m &not specified & $57^\circ\times 42^\circ$&60 &LED & yes/no \\ \hline
DS325  & 320$\times$240 &0.15$-$1 m  &not specified & $74^\circ\times 58^\circ$&60 &diffused laser & yes/no \\ \hline
E70  & 160$\times$120 &0.1$-$10 m & 4 cameras &$70^\circ\times 53^\circ$ &52 &LED & yes/yes \\ \hline
E40   &160$\times$120  &0.1$-$10 m &4 cameras &$45^\circ\times 34^\circ$ &52 &LED & yes/yes \\ \hline
Kinect v2  & 512$\times$424 &0.8$-$4.2 m & not specified &$70^\circ\times 60^\circ$ &30 & LED & yes/no \\ \hline
\end{tabular}
\caption{\label{CW-summary} This table summarizes the main features of commercially available CW TOF cameras. The accuracy of the depth measurements announced by the manufacturers are not reported in this table as the precision of the measurements depend on a lot of factors, such as the surface properties of the scene objects, illumination conditions, frame rate, etc.}
\end{table*}

\section{Calibration of Time-of-Flight Cameras}
\label{section:tof-calibration}

Both pulsed-light and continuous-wave TOF cameras can be modeled as pinhole cameras, using the principles of projective geometry. The basic projection equation is
\begin{equation}
\left(\begin{array}{c}
x\\ y
\end{array}\right)
=
\frac{1}{Z}
\left(
\begin{array}{c}
X\\ Y
\end{array}
\right).
\label{eq:pinhole}
\end{equation}
This implies that the homogeneous coordinates of an image-point $\pvect=(x, y, 1)\tp$ are projectively equal to the scene-coordinates $\Pvect=(X, Y, Z)^\top$, specifically:
\begin{equation}
Z\pvect =
\Pvect.
\label{eq:projection}
\end{equation}
In practice, a realistic model of the projection process involves the intrinsic, extrinsic, and distortion parameters, as described below \cite{zhang-2000,Hartley03,bradski-2008}. 

\subsection{Intrinsic Parameters}
\label{sec:intrinsic-camera}

A digital camera records the image in pixel-units, which are related to the coordinates $(x,y)\tp$ in (\ref{eq:pinhole}) by 
\begin{equation}
\begin{aligned}
u &= \alpha_u x + u_0\\
v &= \alpha_v y + v_0.
\end{aligned}
\label{eq:pixel-coords}
\end{equation}
In these equations we have the following \emph{intrinsic parameters}:
\begin{itemize}
\item Horizontal and vertical factors, $\alpha_u$ and $\alpha_v$, which encode the change of scale, multiplied by the focal length. 
\item The image center, or \emph{principal point}, expressed in pixel units:  $(u_0, v_0)$.
\end{itemize}
%
Alternatively, the intrinsic transformation (\ref{eq:pixel-coords}) can be expressed in matrix form,
$
\qvect
=
\Amat
\pvect
$
where $\qvect=(u, v, 1)$ are pixel coordinates, and the $3\times3$ matrix $\Amat$ is defined as
\begin{equation}
\Amat = \left( \begin{array}{ccc} \alpha_u & 0 &  u_0 \\ 0 & \alpha_v & v_0 \\ 0 & 0 & 1 \end{array} \right).
\end{equation}
Hence it is possible to express the \emph{direction} of a visual ray, in camera coordinates, as
%
\begin{equation}
\pvect = \Amat \inverse \qvect.
\label{eq:direction}
\end{equation}
A TOF camera further allows the 3D \emph{position} of point $\Pvect$ to be estimated, as follows.
Observe from equation (\ref{eq:projection}) that the Euclidean norms of $\Pvect$ and $\pvect$ are proportional:
\begin{equation}
\| \Pvect \| = Z \| \pvect \|.
\end{equation}
The TOF camera measures the distance $d$ from the 3D point $\Pvect$ to the optical center,\footnote{In practice it measures the distance to the image sensor and we assume that the offset between the optical center and the sensor is small}  so
$ d = \| \Pvect \| $.
Hence the $Z$ coordinate of the observed point is
\begin{equation} Z = \frac{\| \Pvect \|}{\| \pvect \|} = \frac{d}{\| \Amat\inverse \qvect\|}. \end{equation}
We can therefore obtain the 3D coordinates of the observed point, by combining (\ref{eq:projection}) and (\ref{eq:direction}), to give
\begin{equation}
\Pvect =  \frac{d}{\| \Amat\inverse \qvect\|} \Amat\inverse \qvect.
\label{eq:xyz-estimate}
\end{equation}
Note that the point is recovered in the camera coordinate system; the transformation to a common `world' coordinate system is explained in the following section.


\subsection{Extrinsic Parameters}
\label{sec:extrinsic-camera}

A rigid transformation from the arbitrary \textit{world coordinate frame} $\Pvect_{\!w} = (X_w, Y_w, Z_w)^\top$ to the camera frame can be modelled by a rotation and translation. This can be expressed in homogeneous coordinates as:
%
%
\begin{equation}
\left( \begin{array}{c} \Pvect \\ 1\end{array} \right) 
=
\left(\begin{array}{cc}
\Rmat & \ \Tvect\\
\mathbf{0}  & 1
\end{array}\right)
\left( \begin{array}{c} \Pvect_{\!w} \\ 1 \end{array} \right).
\label{eq:extrinsic}
\end{equation}
The $3\times3$ matrix $\Rmat$ has three degrees of freedom, which can be identified with the angle and normalized axis of rotation. Meanwhile, the $3\times 1$ translation vector is defined by $\Tvect=-\Rmat\Cvect$, where $\Cvect$ contains the world coordinates of the camera-centre. Hence there are six \emph{extrinsic parameters} in the transformation (\ref{eq:extrinsic}). The equation can readily be inverted, in order to obtain world coordinates from the estimated point $\Pvect$ in equation (\ref{eq:xyz-estimate}).


\begin{figure}[!t]
\centering
\includegraphics[width=0.8\linewidth]{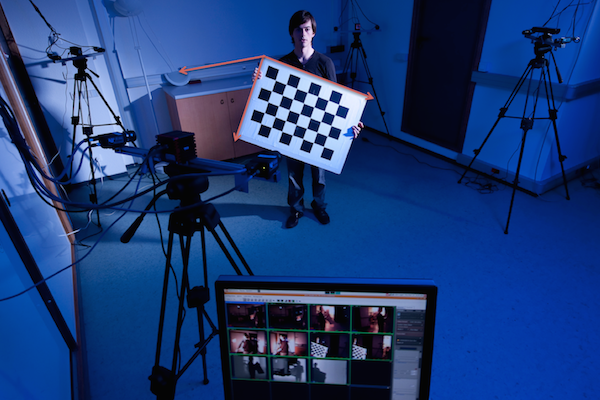} 
\caption{\label{fig:calib-setup} This figure shows a setup for TOF calibration. The calibration board is the same one used for color camera calibration and it can be used to estimate the lens parameters as well.}
\end{figure}

\subsection{Lens Distortion Model}
\label{sec:lens-distortion}
A commonly used lens distortion model widely used for color cameras, \cite{bradski-2008,gonzalez-2011}, can be adopted for \tof{} cameras as well: the observed distorted point $(x_l,y_l)$ results from the displacement of $(x,y)$ according to:
\begin{equation}
\left( \begin{array}{c} x_l \\ y_l \end{array} \right) = 
l_\rho(r) \left( \begin{array}{c} x \\ y \end{array} \right) + \lvect_\tau(x,y)
\end{equation}
where $l_\rho(r)$ is a scalar radial function of
$r = \sqrt{x^2 + y^2}$, and $\lvect_\tau(x,y)$ is a vector tangential component.
These are commonly defined by polynomial functions
\begin{align}
l_\rho(r) & =  1 + \rho_1 r^2 + \rho_2 r^4 \quad\text{and}\\[1ex]
\lvect_\tau(x,y) & = 
\left[
\begin{array}{cc} 
2 x y & r^2 + 2 x^2 \\ 
r^2 + 2y^2 & 2x y 
\end{array} 
\right] 
\left( 
\begin{array}{c} 
\tau_1 \\ \tau_2 \end{array} 
\right)
\label{eqn:lens}
\end{align}
such that the complete parameter-vector is $[\rho_1\; \rho_2\; \tau_1\; \tau_2]$. The images can be undistorted, by numerically inverting (\ref{eqn:lens}), given the lens and intrinsic parameters. The projective linear model, described in sections (\ref{sec:intrinsic-camera}--\ref{sec:extrinsic-camera}), can then be used to describe the complete imaging process, with respect to the undistorted images.

It should also be noted, in the case of TOF cameras, that the outgoing infrared signal is subject to optical effects. In particular, there is a radial attenuation, which results in strong vignetting of the intensity image. This can be modeled by a bivariate polynomial, if a full photometric calibration is required~\cite{lindner-2010,hertzberg2014tof}.

\subsection{Depth Distortion Models}
\label{sec:depth-distortion}
TOF depth estimates are also subject to systematic nonlinear distortions, particulalry due to deviation of the emitted signal from the ideal model described in Section \ref{sec:demodulation}. This results in `wiggling' error of the average distance estimates, with respect to the true distance~\cite{foix-2011,fursattel2015comparative}. 
Because this error is systematic, it can be removed by reference to a precomputed look-up table~\cite{kahlmann2006tof}. \addnote[depth-error]{1}{Another possibility is to learn a mapping between raw depth values, estimated by the sensor, and corrected values. This mapping can be performed using regression techniques applied to carefully calibrated data. A random regression forest is used in \cite{ferstllearning} to optimize the depth measurements supplied by the camera. A kernel regression method based on a Gaussian kernel is used in \cite{kuznetsova2014calibration} to estimate the depth bias at each pixel. Below we describe an efficient approach, 
which exploits the smoothness of the error, and which uses a \mbox{$B$-spline} regression~\cite{lindner-2010} of the form:}
\begin{equation}
d'(x,y) = d(x,y) - \sum_i^n \beta_i B_{i,3}\bigl(d(x,y)\bigr)
\end{equation}
where $d'(x,y)$ is the corrected depth. The spline basis-functions $B_{i,3}(d)$ are located at $n$ evenly-spaced depth control-points $d_i$. The coefficients $\beta_{i}$, $i=1,\ldots,n$ can be estimated by least-squares optimization, given the known target-depths. The total number of coefficients $n$ depends on the number of known depth-planes in the calibration procedure.

\subsection{Practical Considerations}
\label{sec:pract-cons-cal}

\begin{figure*}[t!]
\centering
\includegraphics[width=0.32\linewidth]{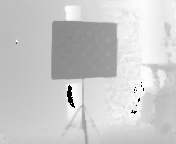}
\includegraphics[width=0.32\linewidth]{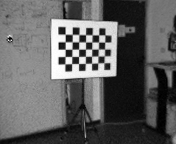} 
\includegraphics[width=.32\linewidth]{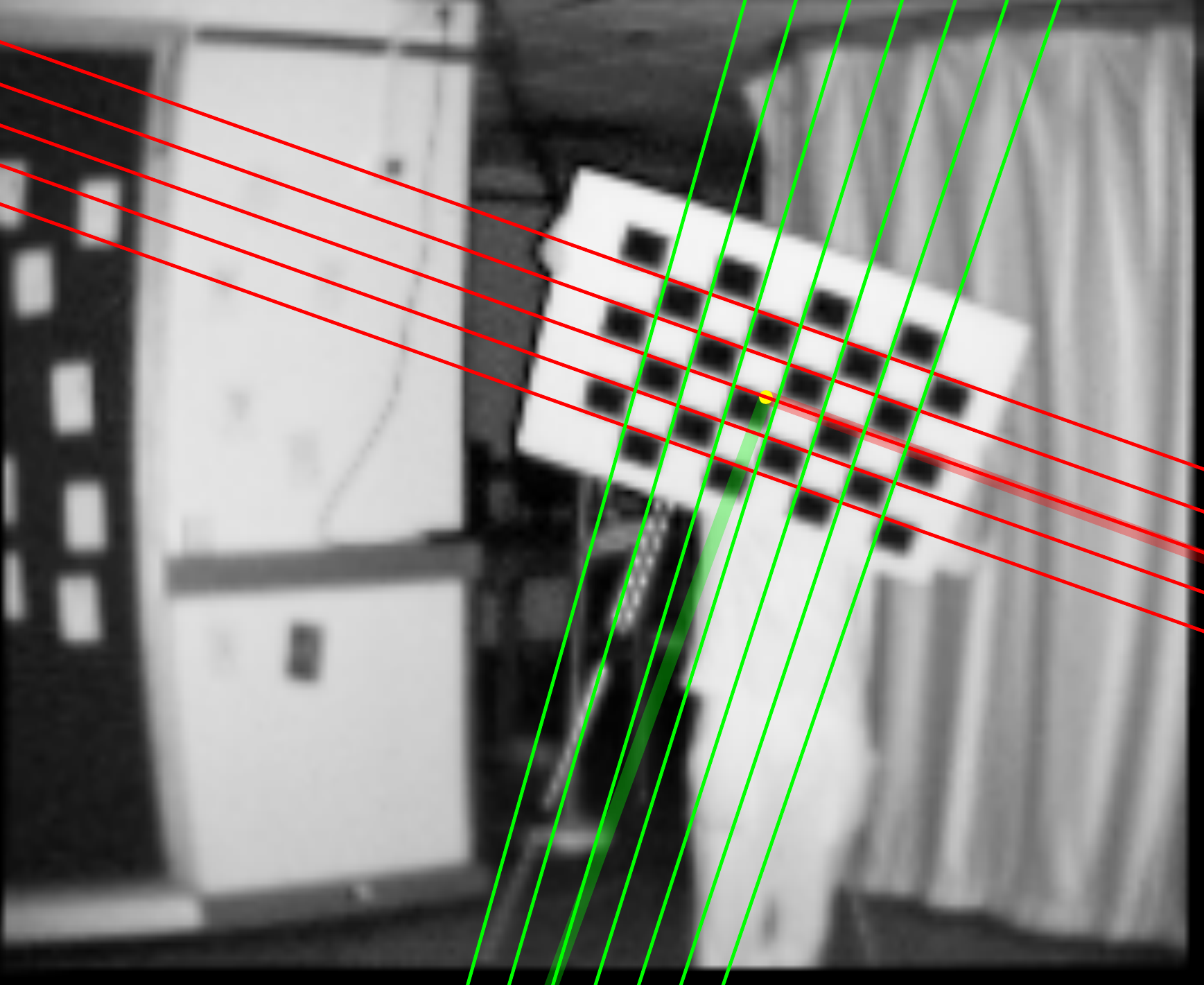}
\caption{\label{fig:TOF-calib} The depth (left) and amplitude (middle) images of an OpenCV calibration pattern grabbed with an SR4000 camera. The actual image resolution is of $176\times 144$ pixels. The amplitude image (middle) has undergone lens undistortion using the model briefly described in section~\ref{sec:lens-distortion}. The calibration pattern detected in the amplitude image (right) using the method of~\cite{Hansard-2014}.}
\end{figure*}

To summarize, the TOF camera parameters are composed of the pinhole camera model, namely the parameters $\alpha_u, \alpha_v, u_0, v_0$ and the lens distortion parameters, namely $\rho_1, \rho_2, \tau_1, \tau_2$. Standard camera calibration methods can be used with TOF cameras, in particular with CW-TOF cameras because they provide an amplitude+offset image, i.e. Section~\ref{section:CW}, together with the depth image: Standard color-camera calibration methods, e.g. OpenCV packages, can be applied to the amplitude+offset image. 
However, the low-resolution of the TOF images implies specific image processing techniques, such as~\cite{Hansard-2014,kuznetsova2014calibration}. As en example, Fig.~\ref{fig:TOF-calib} shows the depth and amplitude images of a calibration pattern gathered with the SR4000 camera. Here the standard corner detection method was replaced with the detection of two pencils of lines that are optimally fitted to the OpenCV calibration pattern. 

TOF-specific calibration procedures can also be performed, such as the depth-wiggling correction \cite{lindner-2010,kuznetsova2014calibration,ferstllearning}. 
A variety of TOF calibration methods can be found in a recent book \cite{Grzegorzek2013time}. A comparative study of several \tof{} cameras based on an error analysis was also proposed \cite{fursattel2015comparative}.

\addnote[multi-path]{1}{One should however be aware of the fact that depth measurement errors may be quite difficult to predict due to the unknown material properties of the sensed objects and to the complexity of the scene. In the case of a scene composed of complex objects, multiple-path distortion may occur, due to the interaction between the emitted light and the scene objects, e.g. the emitted light is backscattered more than once. Techniques for removing multiple-path distortions were recently proposed \cite{freedman2014sra,son2016automatic}.}

\section{Combining Multiple \tof{} and Color Cameras}
\label{section:cross-calibration}
In addition to the image and depth-calibration procedures described in section \ref{section:tof-calibration}, it is often desirable to combine data from multiple devices. There are several reasons for this, depending on the application. Firstly, current depth cameras are pinhole devices, with a single optical centre (\ref{eq:projection}). This means that the resulting point-cloud data is viewpoint dependent; in particular, depth discontinuities will give rise to holes, when the point cloud is viewed from a different direction. Secondly, it is often desirable to combine multiple point clouds, in order to reduce the amount and the anisotropy of sensor noise. Thirdly, typical TOF devices do not capture colour data; hence it may be necessary to cross-calibrate one or more RGB cameras with the depth sensors.

Both pulsed-light and continuous-wave TOF cameras work at a precise wavelength in the (near-)infrared domain. They are equipped with an optical band-pass filter properly tuned onto the wavelength of the light emitter. This allows, in principle, simultaneous capture from multiple TOF cameras with each signaling at its own modulation frequency, so that interference is unlikely.
As mentioned above, not all of the TOF manufacturers allow a user-defined modulation frequency (see Table~\ref{CW-summary}). Because of different spectra, TOF cameras do not interfere with typical color cameras, and can be easily combined. In either combination, however, the cameras must be synchronized properly, that is, a common clock signal should trigger all of the cameras. 

The simplest possible combination is one TOF and one color camera. Some of the available TOF cameras, e.g., Microsoft Kinect~v2 and SoftKinetic  DS311/DS325 have a built-in color camera with its own sensor and lens, which is mounted a few centimeters away from the TOF camera. Note that only real.iZ by Odos Imaging uses the same pixel to acquire both color and depth measurements, thus eliminating the additional calibration and registration. \footnote{There has been an attempt at a similar architecture in~\cite{kim2012cmos}; this 3D and color camera is not commercially available}. 

\begin{figure*}[h!tb]
\centering
\includegraphics[width=0.32\linewidth]{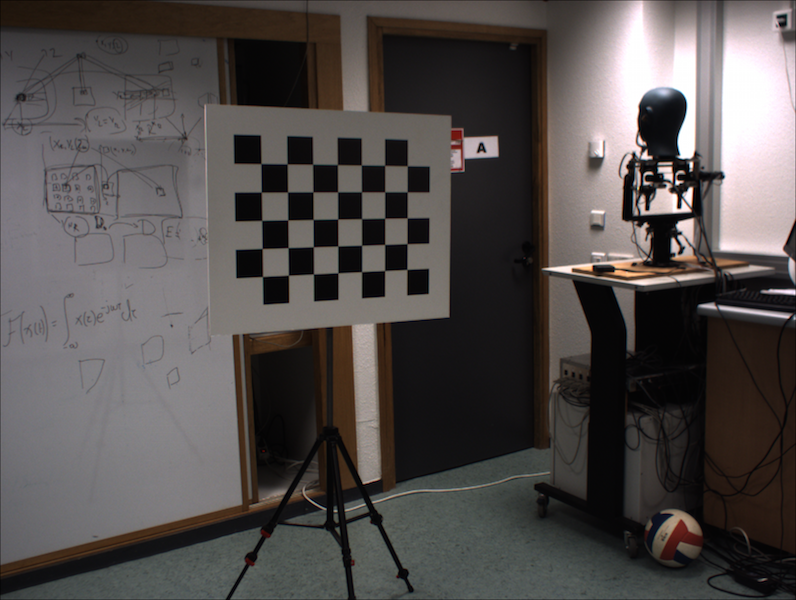}
\includegraphics[width=0.15\linewidth]{tof-depth.png}
\includegraphics[width=0.15\linewidth]{tof-amplitude.png} 
\includegraphics[width=0.32\linewidth]{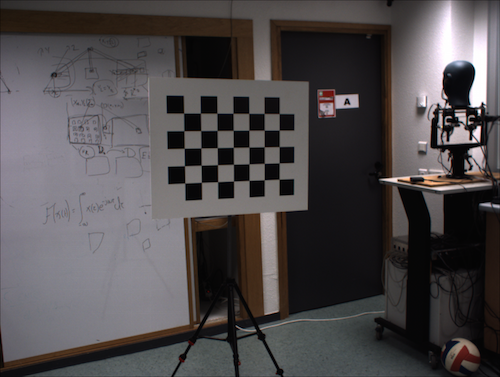}
\caption{\label{fig:TOF-color-ex} Calibration images from synchronized captures. The greyscale images provided by the color camera pair are shown onto the left and onto the right. The middle smaller images correspond to enlarged depth and amplitude images provided by the TOF camera.}
\end{figure*}

 Another possible setup is to use one TOF camera and two color cameras~\cite{Gudmundsson2008,zhu-2008,gandhi-2012, evangelidis-2015,mutto-2015}, e.g., Fig.~\ref{fig:TOF-color-ex}. The advantage of using two color cameras is that they can be used as a stereoscopic camera pair. Such a camera pair, once calibrated, provides dense depth measurements  (via a stereo matching algorithm) when the scene is of sufficient texture and lacks repetitive patterns. However, untextured areas are very common in man-made environments, e.g. walls, and the matching algorithm typically fails to reconstruct such scenes. While TOF cameras have their own limitations (noise, low resolution, etc.) that were discussed above, they provide good estimates regardless of the scene texture. This gives rise to \textit{mixed} systems that combine active-range and the passive-parallax approaches and overcome the limitations of each approach alone. In particular, when a high-resolution 3D map is required, such a mixed system is highly recommended. Roughly speaking, sparse TOF measurements are used as a regularizer of a stereo matching algorithm towards a dense high-resolution depth map~\cite{evangelidis-2015}.
 
Given that TOF cameras can be modeled as pin-hole cameras, one can use multiple-camera geometric models to \textit{cross-calibrate} several TOF cameras or any combination of TOF and color cameras. Recently, an earlier TOF-stereo calibration technique \cite{hansard-2011} was extended to deal with an arbitrary number of cameras \cite{Hansard2015}. It is assumed that a set of calibration vertices can be detected in the TOF images, and back-projected into 3D. The same vertices are reconstructed via stereo-matching, using the high resolution RGB cameras. Ideally, the two 3D reconstructions could be aligned by a rigid 3D transformation; however, this is not true in practice, owing to calibration uncertainty in both the TOF and RGB cameras. Hence a more general alignment, via a 3D projective transformation, was derived. This approach has several advantages, including a straightforward SVD calibration procedure, which can be refined by photogrammetric bundle-adjustment routines. An alternative approach, initialized by the factory calibration of the TOF camera, is described by~\cite{Jung2014}.

A different cross-calibration method can be developed from the constraint that depth points lie on calibration planes, where the latter are also observed by the colour camera~\cite{zhang-2011}. This method, however, does not provide a framework for calibrating the intrinsic (section~\ref{sec:intrinsic-camera}) or lens (section~\ref{sec:lens-distortion}) parameters of the depth camera. A related method, which does include distortion correction, has been demonstrated for Kinect~v1~\cite{herrera-2012}.

Finally, a more object-based approach can be adopted, in which dense overlapping depth-scans are merged together in 3D. This approach, which is ideally suited to handheld (or robotic) scanning, has been applied to both Kinect~v1~\cite{Newcombe11} and TOF cameras~\cite{cui-2013}.

\begin{figure*}[h!tb]
\centering
\includegraphics[height=.3\linewidth]{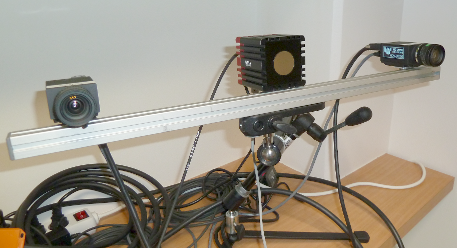}
\includegraphics[height=.3\linewidth]{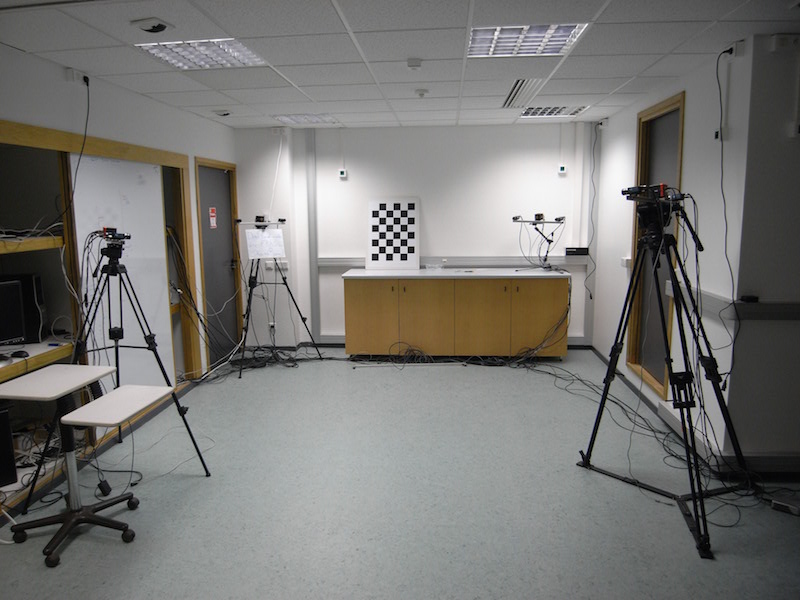} 
\caption{\label{fig:TOF-stereo-setup} A single system (left), comprising a time-of-flight camera in the centre, plus a pair of ordinary color cameras. Several (four) such systems can be combined together and calibrated in order to be used for 3D scene reconstruction (right).}
\end{figure*}

\section{Conclusions}
\label{section:conclusions}
Time of flight is a remote-sensing technology that estimates range (depth) by illuminating an object with a laser or with a photodiode and by measuring the travel time from the emitter to the object and back to the detector. Two technologies are available today, one based on pulsed-light and the second based on continuous-wave modulation. Pulsed-light sensors measure directly the pulse's total trip time and use either rotating mirrors (\lidar{}) or a light diffuser (Flash \lidar{}) to produce a two-dimensional array of range values. Continuous-wave (CW) sensors measure the phase difference between the emitted and received signals and the phase is estimated via demodulation. \lidar{} cameras usually operate outdoor and their range can be up to a few kilometers. CW cameras usually operate indoor and they allow for short-distance measurements only, namely up to 10 meters. Depth estimation based on phase measurement suffer from an intrinsic phase-wrapping ambiguity. Higher the modulation frequency, more accurate the measurement and shorter the range.

Generally speaking, the spatial resolution of these sensors is a few orders of magnitude (10 to 100) less than video cameras. This is mainly due to the need to capture sufficient backscattered light. The depth accuracy depends on multiple factors and can vary from a few centimeters up to several meters. TOF cameras can be modeled as pinhole cameras and therefore one can use standard camera calibration techniques (distortion, intrinsic and extrinsic parameters). \addnote[tof-calib]{1}{One advantage of TOF cameras over depth sensors based on structured light and triangulation, is that the former provides an amplitude+offset image. The amplitude+offset and depth images are gathered by the same and unique sensor, hence one can use camera calibration techniques routinely used with color sensors to calibrate TOF cameras. This is not the case with triangulation-based range sensors for which special-purpose calibration techniques must be used. Moreover, the relative orientation between the infrared camera and the color camera must be estimated as well.}

TOF cameras and TOF range scanners are used for a wide range of applications, from multimedia user interfaces to autonomous vehicle navigation and planetary/space exploration. More precisely:

\begin{itemize}
\item \addnote[lidar]{1}{Pulsed-light devices, such as the Velodyne, Toyota, and Advanced Scientific Concepts \lidar{}s can be used under adverse outdoor lighting conditions, which is not the case with continuous-wave systems. These \lidar{}s are the systems of choice for autonomous vehicle driving and for robot navigation (obstacle, car, and pedestrian detection, road following, etc.). The Toyota \lidar{} scanner is a laboratory prototype, at the time of writing. }

\item The 3D Flash \lidar{} cameras manufactured by Advanced Scientific Concepts have been developed in collaboration with NASA for the purpose of planet landing \cite{amzajerdian2011lidar}. They are commercially available.

\item The SR4000/45000 cameras are used for industrial and for multimedia applications. Although they have limited image resolution, these cameras can be easily combined together into multiple TOF and camera systems. They are commercially available. 

\item The Kinect v2, SoftKinetic, Basler and Fotonic cameras are used for multimedia and robotic applications. They are commercially available, some of them at a very affordable price. \addnote[kinect-lim]{1}{Additionally, some of these sensors integrate an additional color camera which is internally synchronized with the TOF camera, thus yielding RGB-D data. Nevertheless, one shortcoming is that they cannot be easily externally synchronized in order to build multiple-camera TOF-TOF or TOF-color systems.}

\item The real.iZ is a prototype developed by Odos Imaging. It is a very promising camera but its commercial availability is not clear and the time of the writing of this paper.
\end{itemize}
In conclusion, time-of-flight technologies are used in many different configurations, for a wide range of applications. The refinement, commoditization, and miniaturization of these devices is likely to have an increasing impact on everyday life, in the near future.

\bibliographystyle{elsarticle-num}   

\end{document}